%% file: main.tex
\documentclass[letterpaper, 10pt, conference]{IEEEtran}
\usepackage{times}
\usepackage{xr}
\usepackage{amsmath,amssymb,mathrsfs,gensymb}
\usepackage[]{algorithm2e}
\usepackage{graphicx}
\usepackage{xcolor}
\usepackage{wrapfig}
\usepackage{textcomp}
\usepackage{duckuments}

\usepackage[sort,numbers]{natbib}
\usepackage{multicol}
\usepackage[bookmarks=true, hidelinks]{hyperref}

\begin{document}

\title{\vspace{5pt}A non-cubic space-filling modular robot}

\author{Author Names Omitted for Anonymous Review. Paper-ID [add your ID here]}


\author{
\IEEEauthorblockN{%
Tyler Hummer
and
Sam Kriegman}
\IEEEauthorblockA{Northwestern University
}
}

\teaser{
\centering
\includegraphics[trim={0 0 0 0},clip,width=\linewidth]{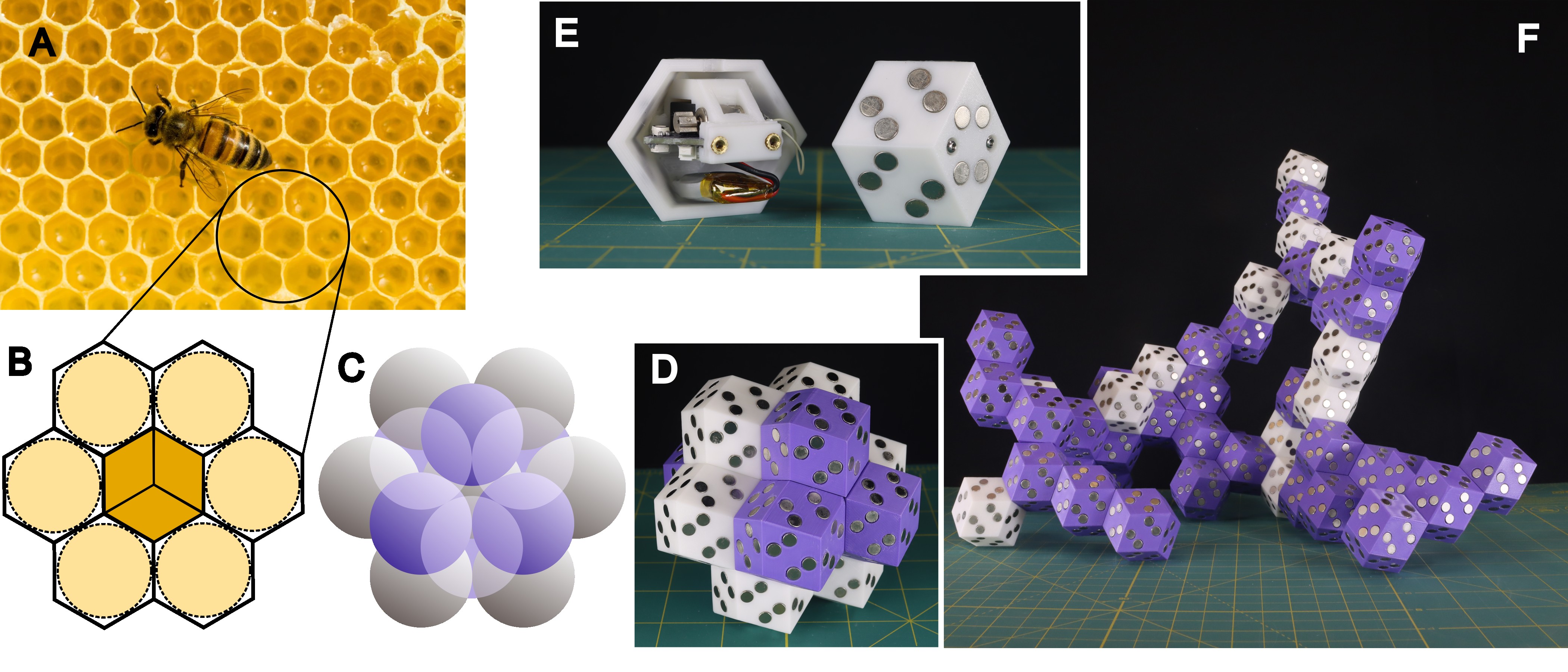} \\
\vspace{-8pt}
\caption{\textbf{Rhombic dodecahedral robots.}
When bees build honeycombs (\textbf{A})
they fashion
three rhombuses at the bottom of each hexagonal chamber (\textbf{B}),
minimizing wax 
and tessellating two layers back-to-back.
Completing this pattern in 3D (\textbf{C}) and closing each cell
yields rhombic dodecahedra (\textbf{D}),
a hitherto nonexistent shape in robotics.
We built dozens of
dodecahedral cells with internal power and actuation (\textbf{E})
that 
attach to form self-motile superstructures (\textbf{F}).
} 
\label{fig:teaser}
\vspace{-24pt}
}

\maketitle

\input{0_abstract}
\IEEEpeerreviewmaketitle

\begin{figure*}
    \centering
    \includegraphics[trim={0 0 0 0},clip,width=\linewidth]{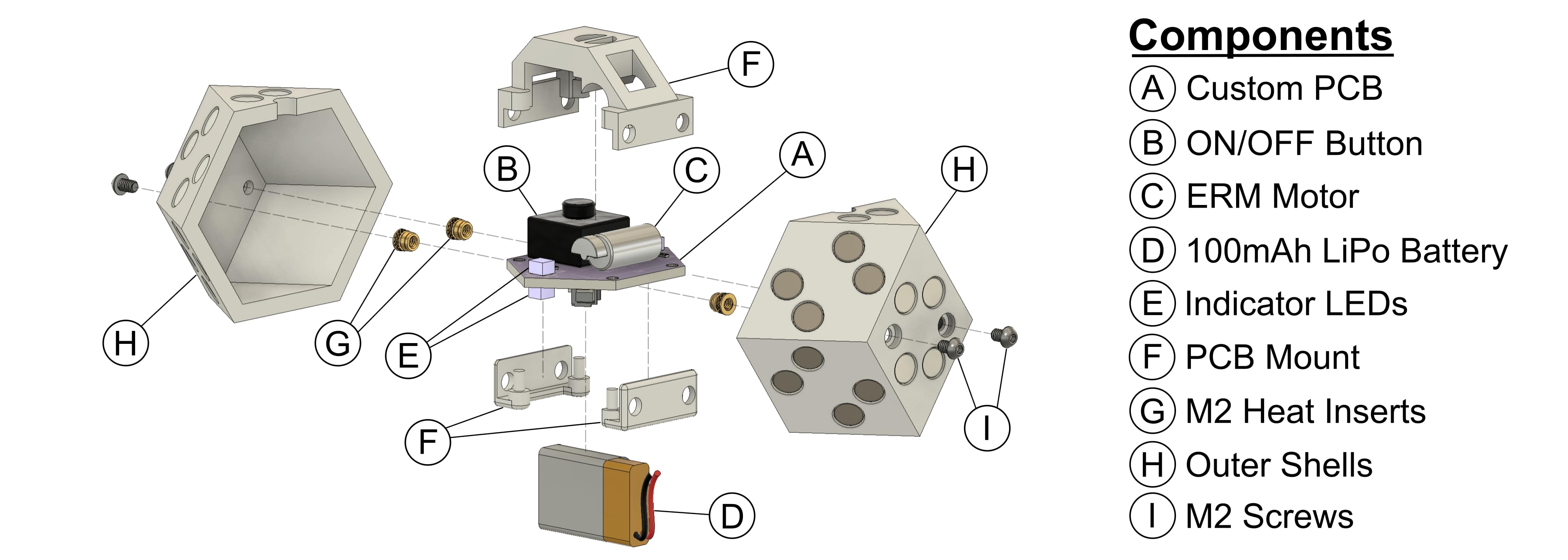}
    \caption{\textbf{Exploded view
    of internal components within an active cell.}
    Active cells include: 
    a hexagonal PCB 
    (\textbf{A}) as a structural foundation and to create all electrical connections; 
    an analog ON/OFF switch (\textbf{B}) for simple mode control;
    an eccentric rotating mass (ERM) motor 
    (\textbf{C}) 
    to provide mechanical energy to the morphological system;
    a 100 mAh LiPo battery cell 
    (\textbf{D}) 
    with high discharge capabilities;
    surface mounted light emitting diodes 
    (LEDs; \textbf{E}) 
    to indicate active cells' current states;
    a 3D printed PCB mount 
    (\textbf{F}) 
    to hold components in place and mount outer shells;
    M2 heat inserts 
    (\textbf{G})
    to mount the outer shells;
    two outer shell halves 
    (\textbf{H}) 
    with 48 face-embedded magnets;
    and M2 screws (\textbf{I}) to complete cell assembly.
    }
    \label{fig:guts}
\end{figure*}

\begin{figure}[b!]
    \centering
    \includegraphics[trim={0 4 0 10},clip,width=\linewidth]{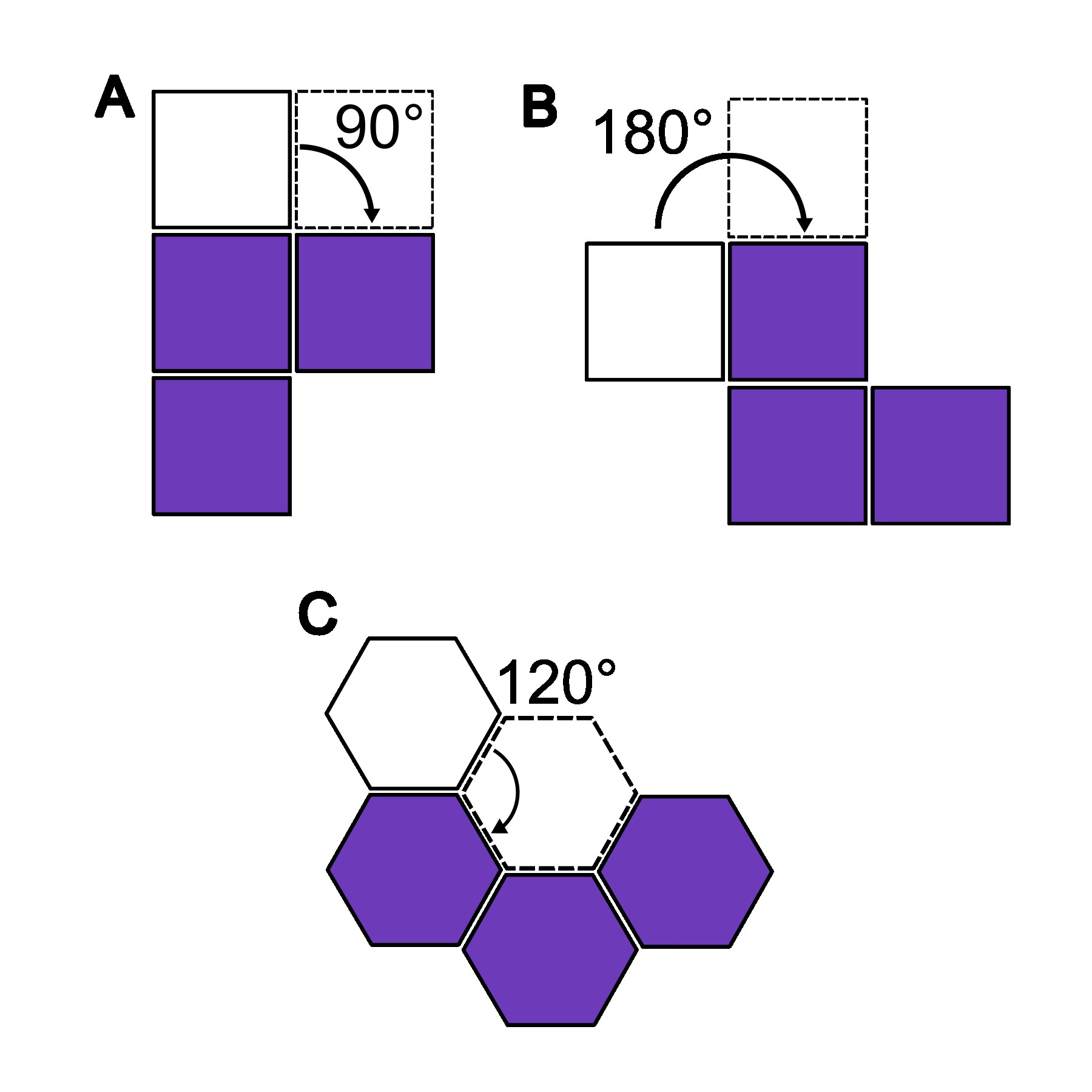}
    \caption{\textbf{The potential for self reconfiguration in future work.} 
    (\textbf{A} and \textbf{B:})
    The two kinds of rolling motions for cubes.
    (\textbf{C:}) 
    2D projection of rhombic dodecahedron rotation (hexagon).
    }
    \label{fig:rotation}
\end{figure}

\input{1_intro}


\input{2_methods}


\begin{figure*}
    \centering
    \includegraphics[width=\linewidth]{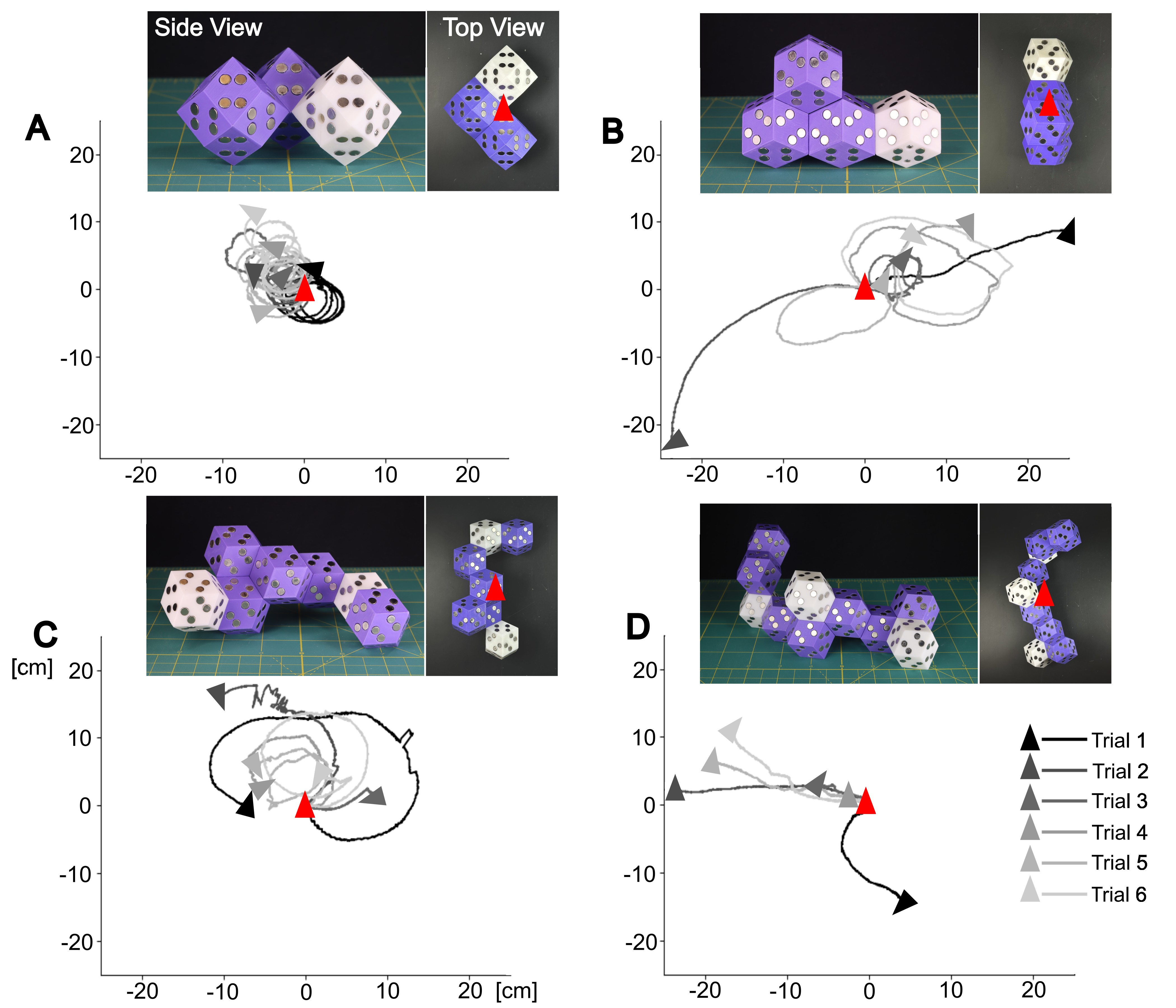}
    \caption{
    \textbf{Locomotion.}
    Behavior was tracked for four different designs (\textbf{A-D}).
    Across six independent trials, design's center of mass was tracked over an evaluation period of 1 min.
    The initial orientation for each trial---%
    the top-down view used for motion tracking---%
    as well as a side view of each morphology, is shown for each respective design. 
    The origin of each trial is marked with a red triangle indicating the initial position and orientation of the robot. 
    The robot's position over the trial period, along with the final position and orientation, are marked with grayscale lines and corresponding triangles.
    Active cells (white) were rotated into six different random orientations, one for each trial, to understand the influence of motor orientation on behavior. 
    In smaller body plans (e.g.~A), the orientation of active cells were found to have greater influence on behavior (e.g.~the direction of rotation) when compared to larger bodies (e.g.~C and D). 
    Potential correlations between the type of surface contact for a given design were also identified.
    For example, point contacts seem to produce a high rate of rotation (A); 
    edge contacts produced large, sweeping rotations (B and C);
    and face contact produced mostly translational motion (D).
    }
    \label{fig:behavior}
\end{figure*}

\input{3_results}


\input{4_discussion}

\input{5_ack}

\bibliographystyle{plainnat}
\bibliography{main}

\end{document}

%% file: 0_abstract.tex
\begin{abstract}
Space-filling building blocks of diverse shape permeate nature at all levels of organization, from atoms to honeycombs, and have proven useful in artificial systems, from molecular containers to clay bricks. But, despite the wide variety of space-filling polyhedra known to mathematics, only the cube has been explored in robotics. Thus, here we roboticize a non-cubic space-filling shape: the rhombic dodecahedron. This geometry offers an appealing alternative to cubes as it greatly simplifies rotational motion of one cell about the edge of another, and increases the number of neighbors each cell can communicate with and hold on to. To better understand the challenges and opportunities of these and other space-filling machines, we manufactured 48 rhombic dodecahedral cells and used them to build various superstructures. We report locomotive ability of some of the structures we built, and discuss the dis/advantages of the different designs we tested. We also introduce a strategy for genderless passive docking of cells that generalizes to any polyhedra with radially symmetrical faces. Future work will allow the cells to freely roll/rotate about one another so that they may realize the full potential of their unique shape.
\end{abstract}

%% file: 1_intro.tex
\section{Introduction}
\label{sec:intro}

When Johannes Kepler peered into the hexagonal cells of honeycombs he saw three rhombi at their base (Fig. \ref{fig:teaser}A,B). 
Capping a cell with three more rhombi, Kepler sketched a rhombic dodecahedron. 
But the shape was already on his mind. 
Studying how snowflakes could be built from microscopic spheres of ice, Kepler made the following conjecture \cite{kepler1611strena}: 
no packing of congruent spheres has density greater than that of the face-centered cubic packing (Fig. \ref{fig:teaser}C), in which each sphere touches 12 others, and which fills just over 74\% of space.%
\footnote{Kepler’s conjecture was formally proven four centuries later, in 2017 \cite{hales2017formal}.}
When uniformly compressed, these spheres will deform into space-filling polyhedra with twelve faces: rhombic dodecahedra (Fig. \ref{fig:teaser}D).

Since Kepler’s discovery of the shape in the early 17th century, rhombic dodecahedra have been found in crystals of 
viruses \cite{glaser1945genes},
proteins \cite{ford1984ferritin}, 
carbohydrates \cite{yang2017giant}, 
metals \cite{niu2009selective} 
and minerals \cite{deer2013minerals}. 
These self-assembling crystalline structures have been of keen interest in several fields of engineering 
due to their exceptional optical, catalytic, electronic
and magnetic
properties 
\cite{wu2011comparative,huang2012synthesis}, 
which suggests that 
space filling in general,
and
the dodecahedral morphology in particular,
may also be useful for self-assembling modular robots \cite{seo2019modular}.

Like crystals, modular robots can ``grow'' to form multicellular bodies by ceaselessly combining and separating unit cells. 
Such robots---if space-filling---could form gapless containers, pipes, mirrors, shields, submersibles, spaceships and extraterrestrial habitats on the fly.
And
by virtue of broader surface contacts between neighbors,
space-filling robots
could build 
much more stable and energetically favorable 
superstructures
than non-space-filling robots \cite{liao2013formation}.

Many unit shapes have been explored in robotics: 
triangles \cite{belke2023morphological}, 
hexagons \cite{pamecha1996design}, 
spheres \cite{swissler2020fireant3d}, 
cubes \cite{romanishin20153d},
cuboids \cite{baca2014modred},
cylinders \cite{li2019particle}
semi-cylinders \cite{salemi2006superbot}
and cylindroids \cite{hirose1996proposal}. 
But only two of these shapes were built to attach in 3D space (spheres and cubes), and only one was capable of filling it: cubes.

The cube is a parallelohedron \cite{senechal1984introduction}: a congruent polyhedron that can be translated to completely fill space, without rotations. 
In other words, space can be partitioned into a regular lattice by cubes. 
In addition to the cube, which has square faces, there are just four other parallelohedra: 
the rhombic dodecahedron (rhombic faces), the hexagonal prism (hexagonal and cubic faces), the elongated dodecahedron (rhombic and octagonal faces), and the truncated octahedron (cubic, rhombic, and hexagonal faces). 
The cube and the rhombic dodecahedron are the simplest parallelohedra. 
Their isometric faces (all squares or all rhombi) allow two cells to connect on all sides, which simplifies the assembly and reconfiguration of multicellular bodies.

There is an infinitude of other polyhedra that, alone or in combination with additional polyhedra, via translations, rotations or reflections, along periodic or aperiodic tilings, can fill space without any gaps. 
Out of all of these possible shapes, only the cube has been taken seriously in robotics. 
Indeed, the vast majority of modular robots built to date have been either cubes or cuboids \cite{liu2016survey}.

Rhombic dodecahedra have an advantage kinematically over cubes that was first noted by \citet{yim1997rhombic}:  
Cubes require two kinds of inherently difficult rolling/rotational motions (90$\degree$ and 180$\degree$; Fig.~\ref{fig:rotation}A and B) to move about each other, and can incur considerable sliding friction against any cubes perpendicular to the axis of rotation.
Rhombic dodecahedra, in contrast, require only a single simple rotational movement (120\degree; Fig.~\ref{fig:rotation}C), and eliminate the need for any part of the cells to slide against each other, while preserving the good property of cubes:
Once two opposing faces of two cells are aligned and connected, 
all rotations about one of the four shared edges will result in another pair of aligned faces.

After neatly illustrating this,
\citet{yim1997rhombic} went on to show how to coordinate the self-assembly of virtual rhombic dodecahedra toward target shapes,
but, until now, no such robots have been created.
Thus here we design and deploy the first rhombic dodecahedra robots.
While they cannot yet assemble themselves, their physicality helps us envisage a future in which they do.

%% file: 2_methods.tex
\section{Methods}
\label{sec:methods}

In this section we detail the design and construction of the robot's cells.

\subsection{Active and passive cells.}

We 3D printed two kinds of rhombic dodecahedra: passive cells (purple) and active cells (white). 
Both active and passive cells consist of a hollow plastic dodecahedral shell (Fig.~\ref{fig:guts}H) with four circular indentations on each face.
Inside each indentation a neodymium magnet was inserted and glued in place (Fig.~\ref{fig:docking}A). 
Passive cells were printed as a single piece of PLA plastic. 

Active cells contain 
a hand-soldered hexagonal printed-circuit-board (PCB), 
an eccentric rotating mass (ERM) vibration motor, 
and a 3.7V 100mAh rechargeable LiPo battery
mounted to the outer shell, which is printed in two parts and screwed together.
An exploded view of an active module is provided in Fig.~\ref{fig:guts}. 
The analog PCB has an ON/OFF toggle switch and LEDs that indicate whether the module is in the ON or OFF state. 
When switched ON, 
the motor rotates at 19000 RPM consuming 195 mA, which allows for a total of $\sim$25 min of continuous operation (approximately half a million revolutions) before depleting its rechargeable battery. 
Each active module costs less than $\$$10 USD to build,
with the most expensive component being the magnets.

\subsection{Docking.}

\begin{figure}[h]
    \centering
    \includegraphics[trim={0 0 0 0},clip,width=\columnwidth]{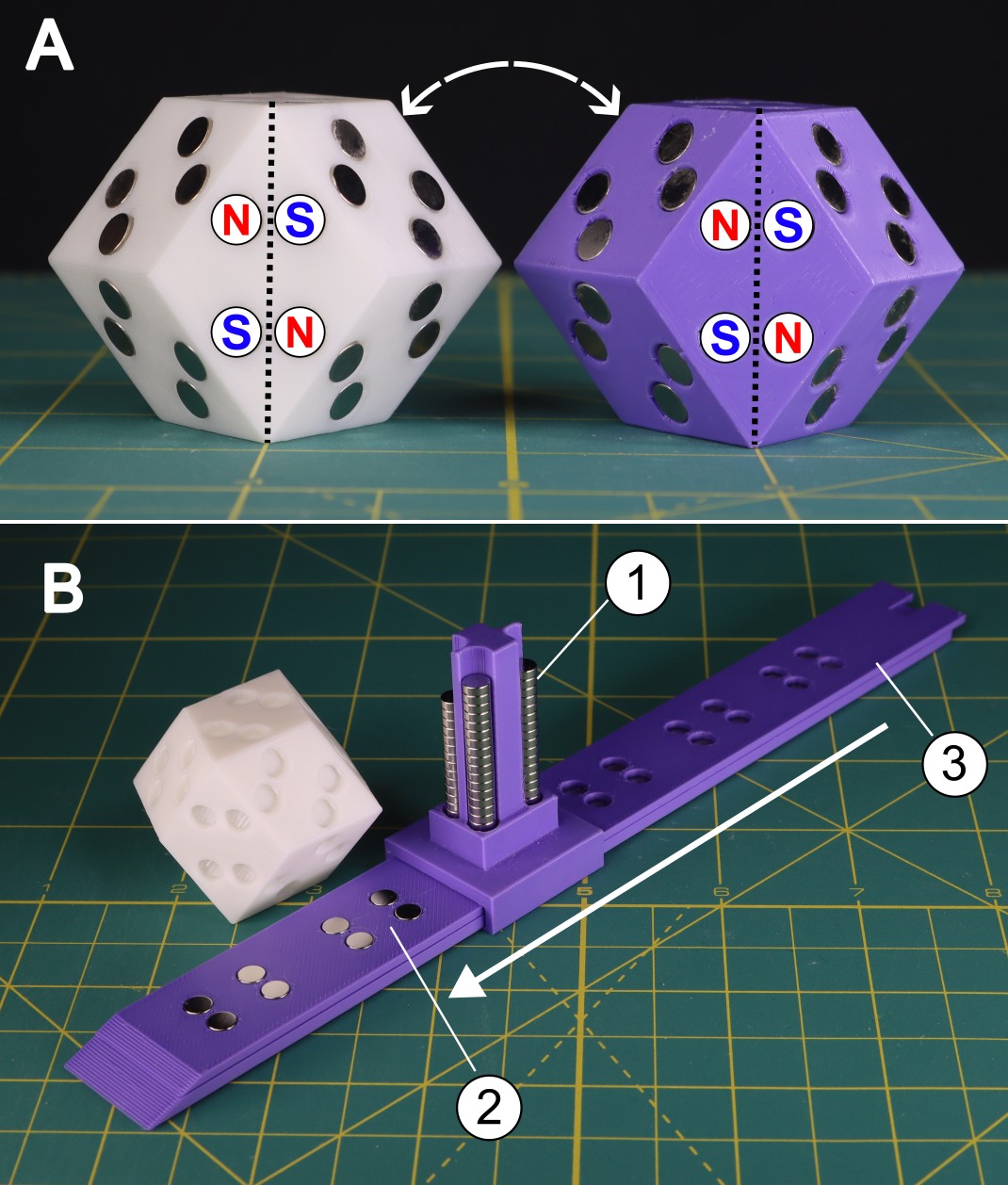}
    \caption{\textbf{Genderless passive docking} was achieved by arranging and embedding axially-poled neodymium magnets into the outer shell of the cells (\textbf{A}). 
    The arrangement of north (N) and south (S) poled magnets takes advantage of the symmetry line that exists along the long axis of the rhombic faces. 
    (\textbf{B:}) A jig was designed and 3D printed to simplify cell assembly. 
    The jig consists of a base (B1) to hold and supply magnets, a slider base (B2) with identical magnet configurations as shown in A 
    that automatically loads itself when slid through the base, and a slider top (B3) that can be removed to allow all four magnets in a cell face to be embedded simultaneously.
    }
    \label{fig:docking}
\end{figure}

Genderless passive magnetic docking allows any two cells to dock at any aligned face, without rotations (Fig.~\ref{fig:docking}A), in a fashion similar to that of biological cells \cite{kriegman2020xenobots,kriegman2021kinematic}.

Initial prototypes incorporated diametrically-poled neodymium magnets that could freely rotate into the necessary orientation to enable cell-cell docking. 
However, the scale and shape of the cells caused the orientation of magnets to lock pole-to-pole within the cell, which prevented the magnets from rotating
and thus inhibited the cell from docking with another.
To rectify this, four axially-poled neodymium magnets were embedded on each of the cell's 12 faces 
(48 total magnets per cell). 

There are many ways to arrange these magnets to ensure a stable connection.
For example, 
magnets could be placed in the four corners of each face.
However, this would not always allow for two geometrically aligned cells to connect magnetically.
While a new cell could always be rotated to connect to the end of a 1D chain,
in 2D and 3D structures a valid rotation will not always be possible, rendering connections unstable (one or more pairs of magnets are repulsive) or impossible (no pairs of magnets are attractive). 

\begin{figure*}
    \centering
    \includegraphics[width=\linewidth]{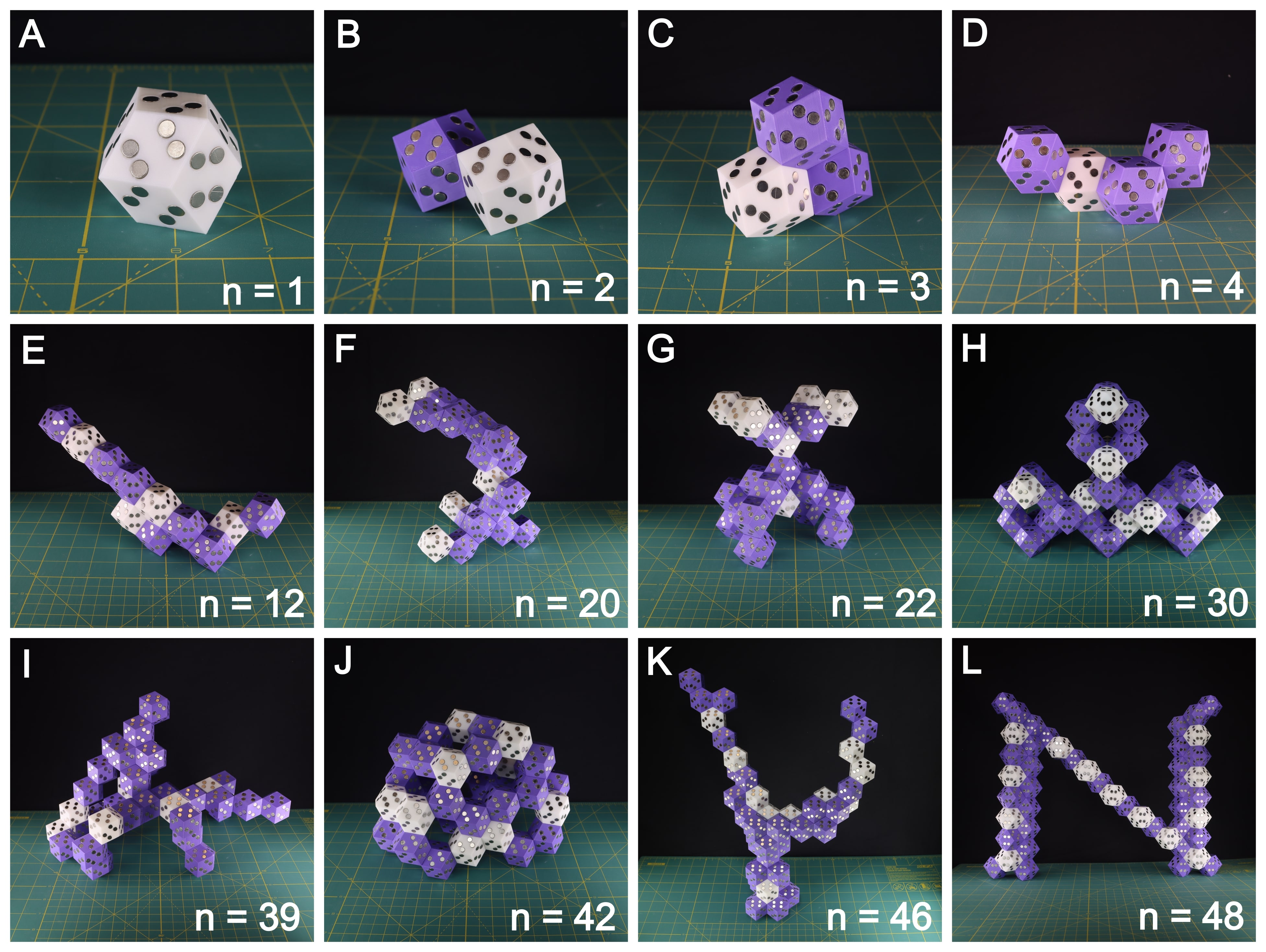}
    \caption{\textbf{Robotic crystals}
    formed by various numbers (n) of space-filling unit cells.} 
    \label{fig:gallery}
\end{figure*}

To ensure genderless docking, 
pairs of magnets along the two opposing faces of two aligned cells must preserve their unlike poles under 180$\degree$ rotations.
The magnet configuration we chose respects this constraint (Fig.~\ref{fig:docking}A).
With this magnet configuration, cells can connect within the entire face-centered cubic lattice-configuration that is characteristic of rhombic dodecahedra (Fig.~\ref{fig:teaser}C).
This provides
orientation-invariant and configuration-invariant attachment of cells, or groups of cells (Fig.~\ref{fig:gallery}),
and is not limited to rhombic dodecahedra. 
Indeed, this strategy generalizes to any polyhedra with faces 
that possess 
at least two degrees of rotational symmetry.
Examples of such shapes include the regular icosahedron, which has 20 equilateral triangular faces with three degrees of rotational symmetry on each face; 
the gyrobifastigium, which has four equilateral triangular faces and four square faces (four degrees of rotational symmetry); 
and the truncated octahedron, which has eight hexagonal faces (six degrees of rotational symmetry) and six square faces.

Despite the simplicity that passive magnetic docking enables, complications still arose. 
The first was that,
due to the magnet's smooth surface finish, glue could not properly adhere to their surface, and 
as a result, embedded magnets were prone to falling out.
This problem was ameliorated by sanding down
the surface finish of the magnets to increase their roughness and surface friction prior to mounting.
This greatly reduced the number of magnets that dislodged from the face during reconfiguration.


Another challenge was that embedding magnets manually, one by one, is a time consuming and labor intensive task that required upwards of 30 min per cell.
We addressed this problem by designing a jig to hold the magnets and guide them into place (Fig.~\ref{fig:docking}B). 
The jig consists of only three parts: 
a base that holds and supplies the magnets, 
a slider that has identical magnet configurations as the cells, 
and a removable slider top. 
The slider is moved through the slot in the base allowing the magnets to reload themselves into the desired location. 
The slider top can then be removed and the entire face of a cell can be mounted simultaneously.  
With these refinements, one cell can be embedded with magnets in 10 min, and is now primarily limited by the curing time of the glue used for mounting.

%% file: 3_results.tex
\section{Results}
\label{sec:results}

Four unique morphologies comprising active and passive cells were configured manually and tested for their locomotive abilities (Fig.~\ref{fig:behavior}). 
Six trials were recorded, each lasting 1 min. 
At the beginning of each trial,
the body was placed at a marked central location with the same initial orientation. 
To isolate the influence of 
motor orientation 
from that of
body shape and motor position, the
active cells within each design were rotated into a different random orientation for each trial, changing the orientation and directionality of the ERM motor within them.
Although an exhaustive test of motor orientations would be ideal, given that each active cell has 12 sides, each of which can be attached in two ways, 
testing each motor orientation of each active cell was not feasible under the given time constraints.

Motion of the bodies was tracked using Blender%
\footnote{\href{https://www.blender.org}{\color{blue}www.blender.org}
}
and extracted for analysis.%
\footnote{\href{https://github.com/Amudtogal/blenderMotionExport}{\color{blue}github.com/Amudtogal/blenderMotionExport}
}
Tracking boxes within Blender were placed along all of the upwards facing cells of the body, and then joined to track the center of mass. 
One advantage of the chosen magnet arrangement is that it offers a plethora of markers for motion tracking without needing to add sticker tags. 
The mirror finish of the magnets can cause issues with light reflections at certain orientation, 
but the effects of this can be minimized with a sufficient number of tracking markers.


With these four unique morphologies, 
four distinct behaviors emerged. 
Design A (Fig.~\ref{fig:behavior}A) was the smallest morphology we tested, comprising just three cells: two passive and one active.
It was extremely motile, traveling a mean distance of 126 cm $\pm$ 34 SD over the six trials; 
however, the motion was largely rotational rather than translational, resulting in a mean net displacement of only 6 cm $\pm$ 4 SD.
The direction of rotation, clockwise or counterclockwise, was greatly influenced by the orientation of the motor within the passive cell.
Out of the six trials, 66.67\% random rotations of the single active cell caused Design A to turn clockwise, and in the other 33.33\% it turned counterclockwise.

Design B (Fig.~\ref{fig:behavior}B) is also quite small (three passive, one active) and is also very sensitive to motor orientation.
Exactly half of the random rotations applied to its single active cell caused Design B to turn clockwise, and in the other half of the trials it turned counterclockwise.

\begin{table*}[!ht]
    \normalsize
    \centering
    \label{table}
    \caption{Morphological and behavioral characterization of the four designs we tracked.}
    \begin{tabular}{rcccc}
        \textbf{} & \textbf{Design A} & \textbf{Design B} & \textbf{Design C} & \textbf{Design D } \\ \hline
        No. of passive cells  & 2 & 3 & 5 & 7  \\ 
        No. of active cells  & 1 & 1 & 2 & 3  \\ 
        Ratio of passive to active  & 2 to 1 & 3 to 1 & 2.5 to 1 & 2.33 to 1  \\ 
        Body length (cm)  & 9.5 & 11.5 & 15 & 21  \\ 
        Body weight (g)  & 77 & 98 & 175 & 252  \\ 
        Type of surface contacts  & Point  & Edge & Edge & Face  \\ 
        Avg. distance traveled (cm)  & 126 +/- 34 SD & 76 +/- 14 SD & 106 +/- 22 SD & 79 +/- 26 SD  \\ 
        Avg. net displacement (cm)  & 6 +/- 4 SD & 22 +/- 21 SD & 9 +/- 5 SD & 16 +/- 10 SD  \\ 
        No. of trials  & 6 & 6 & 6 & 6 \\ \hline
    \end{tabular}
\end{table*}

Design C (Fig.~\ref{fig:behavior}C) comprised five passive cells and two active cells.
Similar to Design A, 
Design C was highly motile, traveling an average of 106 cm $\pm$ 22 SD across the six trials.
However, contrary to the inconsistent direction of rotation exhibited by Design A, 
Design C consistently traveled in a counterclockwise direction of rotation in all six of the trials. 
This may imply that as the number of cells within a given morphology grows, the orientation of the motor within the active cells has less influence on the overall behavior.
Thus, the behavior could become increasingly a result of the \textit{position} of active and passive cells. 

This may also be supported by the mostly translational motion of Design D (Fig.~\ref{fig:behavior}C), which exhibited a mean distance of travel of 79 cm $\pm$ 26 SD, and a mean net displacement of 16 cm $\pm$ 10 SD.
With seven passive and three active cells, Design D was the largest of the four designs we tested.
In the majority of trials (66.67\%) Design D moved 
translationally.

Another characteristic of interest is the type of surface contact that a given morphology maintains with the ground plane. 
Due to the geometrical properties of the rhombic dodecahedron, all cells with a given morphology have identical orientations. 
As a result, each design can have only a single type of surface contact: point, edge, or face. 
Each type of contact was represented by at least one of the tested designs. 

Design A maintained point contact with the ground, and also exemplified the largest rate of consistent rotation, regardless of direction.
Designs B and C both maintained edge contact. 
Their behavior, as with Design A, was characteristically rotational, but with larger, sweeping paths. 
Design D, on the other hand, maintained face contact with the ground plane and displayed more consistent translational motion. 
These results suggest that the type of surface contact
not only influences
the emergent behavior of a given design,
but that surface contacts could be intentionally modified to achieve desired behaviors.

%% file: 4_discussion.tex
\section{Discussion}
\label{sec:discussion}

Inspired by 
crystals across length scales, as well as sheer intellectual curiosity,
we set out to build a non-cubic space-filling robot.
We zeroed in on the rhombic dodecahedron rather quickly because it seemed to us to be the simplest option.
We later discovered that we were not the first to dream of robots with this shape.
Three decades ago, \citet{yim1997rhombic} clearly illustrated the issues with self-assembling cubes and how the rhombic dodecahedron obviates them.
Lost to history, no such robots were ever built, nor have they been discussed since.

Here we demonstrated the first robots composed of cells with rhombic dodecahedron shape.
At the moment these self-motile machines are 
more likely to 
raise debates 
than carry out useful work.
What must a machine do, in addition to moving around, 
to be considered a bona fide robot?
Future work could closely follow the cadence of this argument, 
squeezing sensing, computation and communication inside and along the shell of each module.
But, so long as the individual cells cannot intentionally detach, roll themselves over 120$\degree$
and reattach, these robots will not be able to do the very thing that, at least according to \citet{yim1997rhombic}, makes them so~special.

To get the cell rolling, 
internal mechanisms 
already used by self-assembling cubes, 
such as a flywheel and brakes
\cite{romanishin20153d},
could be adapted for six axes of rotation instead of three.
But this will likely require a considerable investment in terms of time, money and effort.
Future cells could also be built with materials that support active de/magnetization
and 
respond to external forces such as light and rotating magnetic fields 
\cite{li2020fast}
to allow the cells to passively
roll up and down the robot's body.
Existing computationally efficient topology search algorithms \cite{matthews2023efficient} could be adapted for internally modeling the repercussions of such actions prior to reconfiguration.

That said, we would like to emphasize  
the value of designing, building and playing with this first generation of non-rolling dodecahedral cells.
The first hand experience of
manipulating physical rhombic dodecahedra---%
manually rotating and snapping them together in different arrangements---%
helped us
better understand 
the mathematics of crystals 
and
fully grasp
the kinematics of robots with dodecahedral morphology.
We also found playing with them to be quite fun.
They provide a satisfying tactile experience due to 
their symmetry, 
obtuse angles which let them roll in your hand, 
and the building tension of twelve magnetic faces.
This suggests that they may provide an engaging and interactive educational tool for learning about geometry, crystals and robots.

%% file: 5_ack.tex
\section*{Acknowledgements}

We thank Elaine Liu for help designing early prototypes, 
Davin Landry for help with printing and process development for cell assembly,
and Stefan Knapik, David Matthews, 
and other members of the Center for Robotics and Biosystems at Northwestern University 
for helpful discussions.
This research was supported by
a seed grant from the Center for Engineering Sustainability and Resilience at Northwestern University.